\documentclass[preprint,12pt]{elsarticle}

\usepackage{listings}
\usepackage{color}

\definecolor{dkgreen}{rgb}{0,0.6,0}
\definecolor{gray}{rgb}{0.5,0.5,0.5}
\definecolor{mauve}{rgb}{0.58,0,0.82}

\usepackage{tikz}
\usetikzlibrary{backgrounds}
\usepackage{graphicx}
\usepackage{bm}
\usepackage{amsmath}
\usepackage{setspace}
\usepackage{hyperref}
\usepackage{subfigure}
\usepackage{multirow}
\hypersetup{
    colorlinks=true,       % false: boxed links; true: colored links
}
\usepackage{amssymb}
\usepackage{lineno}
\usepackage{diagbox}

\biboptions{sort&compress}

\journal{Journal Name}

\begin{document}

\begin{frontmatter}


\title{Open Problems in Applied Deep Learning}



% \author{Maziar Raissi$^{1}$, Paris Perdikaris$^{2}$, and George Em Karniadakis$^{1}$}
% \address{$^{1}$Division of Applied Mathematics, Brown University,\\ Providence, RI, 02912, USA\\
% $^{2}$Department of Mechanical Engineering and Applied Mechanics,\\ University of Pennsylvania,\\ Philadelphia, PA, 19104, USA}

\author{Maziar Raissi}
% \address{Division of Applied Mathematics, Brown University,\\ Providence, RI, 02912, USA}
\address{Department of Applied Mathematics, University of Colorado Boulder,\\ Boulder, Colorado, 80309, USA}

\begin{abstract}
This work formulates the machine learning mechanism as a bi-level optimization problem. The inner level optimization loop entails minimizing a properly chosen loss function evaluated on the training data. This is nothing but the well-studied training process in pursuit of optimal model parameters. The outer level optimization loop is less well-studied and involves maximizing a properly chosen performance metric evaluated on the validation data. This is what we call the ``iteration process'', pursuing optimal model hyper-parameters. Among many other degrees of freedom, this process entails model engineering (e.g., neural network architecture design) and management, experiment tracking, dataset versioning and augmentation. The iteration process could be automated via Automatic Machine Learning (AutoML) or left to the intuitions of machine learning students, engineers, and researchers. Regardless of the route we take, there is a need to reduce the computational cost of the iteration step and as a direct consequence reduce the carbon footprint of developing artificial intelligence algorithms. Despite the clean and unified mathematical formulation of the iteration step as a bi-level optimization problem, its solutions are case specific and complex. This work will consider such cases while increasing the level of complexity from supervised learning to semi-supervised, self-supervised, unsupervised, few-shot, federated, reinforcement, and physics-informed learning. As a consequence of this exercise, this proposal surfaces a plethora of open problems in the field, many of which can be addressed in parallel.
\end{abstract}

\begin{keyword}
supervised learning, semi-supervised learning, self-supervised learning, unsupervised learning, few-shot learning, federated learning, reinforcement learning, and physics-informed learning
\end{keyword}

\end{frontmatter}

% \linenumbers

\section{Introduction}
\begin{figure}[htbp]
\centering
\includegraphics[width=\textwidth]{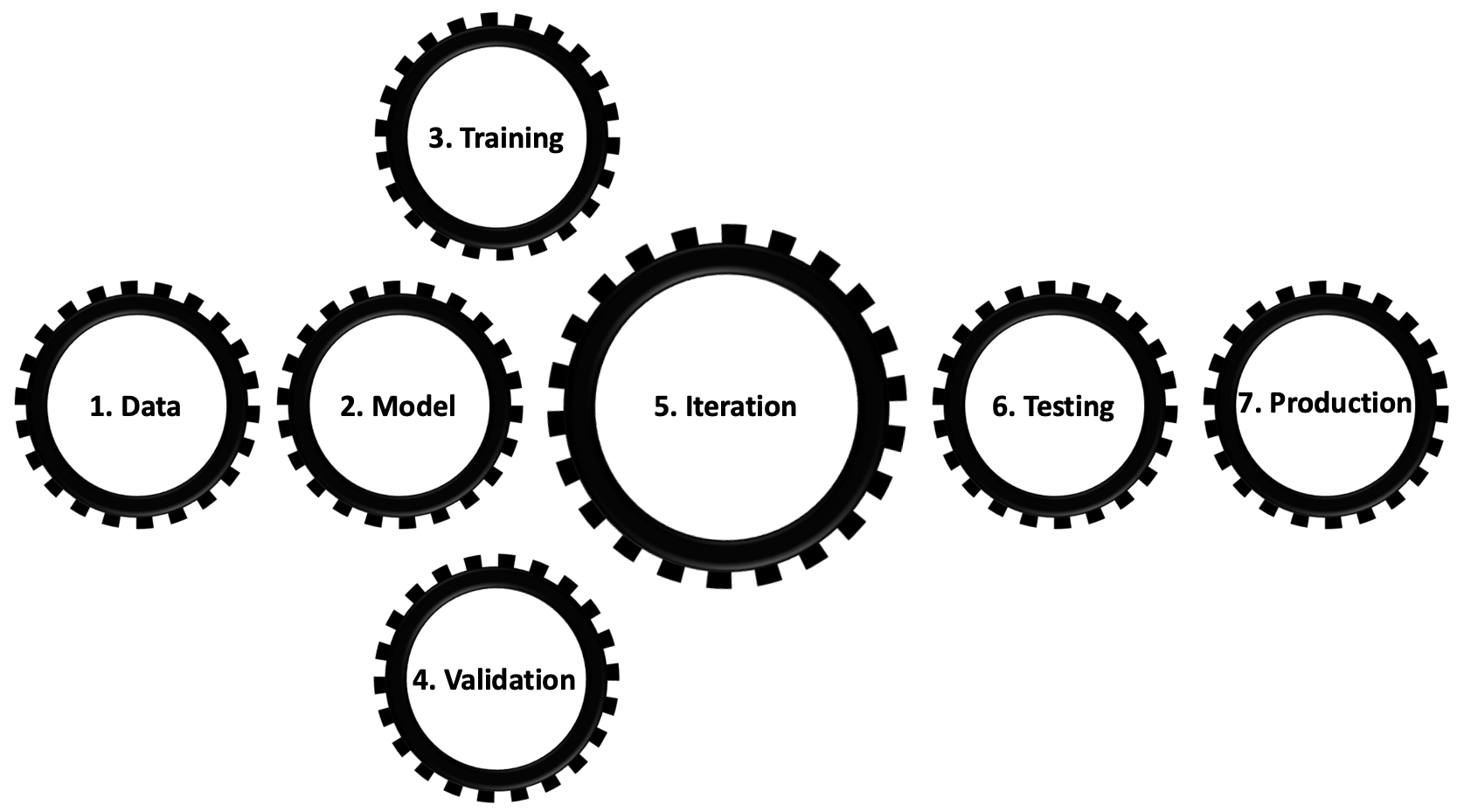}
\caption{The general mechanism for building machine learning solutions.}
\label{fig:big_picture}
\end{figure}

The general mechanism for building machine learning solutions is illustrated in Fig. \ref{fig:big_picture} and outlined in the following. 1) Everything starts with data as the ``source code'' for machine learning. 2) We would then write a model to fit the data. 3) The model is then trained to maximize the likelihood of the training data or minimize a distance/divergence between the training data distribution and model predictions in the case of generative adversarial networks. The training process typically entails multiple steps of stochastic gradient descents (e.g., Adam optimizer \cite{kingma2014adam}). 4) The model is then evaluated using a properly chosen performance metric (e.g., accuracy, mean average precision, inception score, etc.) on the validation data. 5) Next is the iteration step where the aforementioned steps 1-4 need to be repeated tens of thousands of times to find the most performant solution. This step entails model engineering (e.g., neural network architecture design) and management, experiment tracking, dataset versioning/augmentation, in addition to seemingly ``minor'' details such as choosing learning rates and learning rate schedules, batch sizes, weight decay and batch/weight/layer/group/spectral normalizations, just to name a few. The iteration step is a crucial piece of the machine learning pipeline and is usually the most time and resource consuming step while often being overlooked. This is what makes machine learning difficult. 6) Before putting the model into production, we test it one last time on some test data. 7) The final stage is serving the model in production to millions of customers/users while constantly monitoring its performance and re-training it if needed.\\

Step 5 (i.e., the iteration step) is the topic of this work. The jury is still out whether the iteration step should be automated (i.e., AutoML) or left to the intuitions of machine learning students, engineers, and researchers. Regardless of the route we take, there is a need to reduce the computational cost of the iteration step and as a direct consequence reduce the carbon footprint of developing artificial intelligence algorithms \cite{strubell2019energy}. The computational bottleneck of the iteration step is the training of each model to convergence, only to measure its performance on the validation data which is then used as the feedback signal to guide the iteration process.

\section{Background}\label{sec:background}
In mathematical notations, we are trying to solve bi-level optimization problems \cite{anandalingam1992hierarchical, colson2007overview} of the following form;
\begin{equation}
\begin{aligned}
\min_{\alpha} \quad & \mathcal{M}_{\text{val}}(w^*(\alpha), \alpha)\\
\textrm{s.t.} \quad & w^*(\alpha) = \arg \min_w \mathcal{L}_{\text{train}}(w,\alpha).\\
\end{aligned}
\label{eq:multi_task}
\end{equation}
Here, $\alpha$ denotes the ``hyper-parameters'' of the model in an abstract sense, encapsulating concepts as generic as learning rate (schedule), depth and width of neural networks, discrete choices between CNNs, RNNs, or Transformers \cite{vaswani2017attention} and their variants \cite{dosovitskiy2020image}, presence or absence of different types of normalization layers \cite{wu2018group}, to pad or not to pad, convolutional kernel sizes, to use dropout or not, etc. Moreover, $\mathcal{M}_{\text{val}}$ denotes a performance metric such as accuracy, mean average precision \cite{padilla2020survey}, Fréchet Inception Distance \cite{heusel2017gans}, etc. The loss function is denoted by $\mathcal{L}_{\text{train}}$ while the model parameters are represented by $w$. Whereas Eq. \ref{eq:multi_task} explains the iteration problem (see Fig. \ref{fig:big_picture}) in a clean and unified mathematical formulation, the solution to this problem is complex and case specific.\\

For some machine learning problems, the evaluation metric $\mathcal{M}_{\text{val}}$ is a discrete and non-differentiable function of $\alpha$ (e.g., accuracy \cite{krizhevsky2012imagenet}). For some other problems, it isn't well-defined (e.g., domain adaptation \cite{ganin2016domain}, semi-supervised \cite{xie2020self}, self-supervised and unsupervised learning \cite{grill2020bootstrap, chen2020simple} as well as generative models \cite{salimans2016improved}) or is even non-existent (e.g., \href{https://www.youtube.com/watch?v=GXZq2_WYRjo}{physics-informed deep learning} \cite{pinns_2019, raissi2019physics, Raissi2018DeepHP, Raissi2020HiddenFM}). The hyper-parameters $\alpha$ could be continuous (learning rate, momentum, weight-decay, etc.) or discrete variables (e.g., the choice between a regular convolution or a depth-wise-separable one \cite{chollet2017xception, howard2017mobilenets}, number of layers/channels, to use batch-norm \cite{ioffe2015batch} or not, etc.). Sometimes $\alpha$ is a function of the current training epoch (e.g., learning rate schedule, progressive growing of GANs \cite{karras2017progressive}, etc.). More often than not, the loss function is a weighted combination of multiple loss functions (e.g., multi-task learning \cite{isola2017image}, physics-informed machine learning \cite{raissi2018deep}, etc.); Those weights could be part of $\alpha$ and require special treatment to balance the trade-off between focusing on one objective function versus another. The loss function $\mathcal{L}_{\text{train}}$ could itself be discrete and non-differentiable (e.g., cost/reward in reinforcement learning \cite{andrychowicz2020learning}). Therefore, the solutions to the problem specified in Eq. \ref{eq:multi_task} and illustrated in Fig. \ref{fig:big_picture} are not as clean as its formulation and end up being case specific. The first solution that comes to mind is to rely on the intuition of students, researchers, and engineers obtained through tens of thousands of collective trial and errors and knowledge sharing over the Internet (e.g., stackoverflow.com), blog posts, and social media platforms. We can also build tools to facilitate such experimentations. A good example is ``\href{https://wandb.ai/site}{Weights \& Biases}'', an MLOps platform to help researchers build better models faster with experiment tracking, dataset versioning, and model management. MLOps is an active area of research both in academia and industry. In the cases where $\mathcal{M}_{\text{val}}$ is a well-defined function, let it be discrete or non-differentiable, one could use grid search (albeit for low dimensional $\alpha$), random search \cite{cubuk2020randaugment}, Bayesian optimization \cite{kandasamy2018neural}, reinforcement learning \cite{zoph2016neural, zoph2018learning, cubuk2019autoaugment} or evolutionary algorithms \cite{real2019regularized, miikkulainen2019evolving} to find $\alpha^*$, even if $\alpha$ is discrete. The shortcoming of such approaches for solving Eq. \ref{eq:multi_task} is their extensive computational cost (e.g., 32,400-43,200 GPU-Hours, the equivalent of using 450 GPUs for 3-4 days \cite{pham2018efficient, liu2018darts}). Such approaches have therefore a high carbon footprint \cite{strubell2019energy}. The bottleneck is solving the inner optimization problem in Eq. \ref{eq:multi_task} (i.e., the training loop) to completion, only to measure the model's performance on the validation data which is then used as the feedback signal to guide the iteration process (i.e., the outer optimization problem in Eq. \ref{eq:multi_task}). One solution to this problem is to trade off computation for more memory consumption using parameter sharing ideas \cite{pham2018efficient}. This could result in up to 1000x faster solutions to Eq. \ref{eq:multi_task} (e.g., 16 GPU-Hours) in some cases. The idea is to warm-start the training process using some shared parameters cached in memory, rather than starting from random parameters (i.e., cold-starting). This could speed up the convergence of the inner optimization problem in Eq. \ref{eq:multi_task}. Alternatively, one could approximate the validation function $\mathcal{M}_{\text{val}}$ with a differentiable function (e.g., the loss function on the validation data $\mathcal{L}_{\text{val}}$ \cite{liu2018darts}). Doing so will enable us to use plain-vanilla stochastic gradient descent algorithms (e.g., the Adam optimizer \cite{kingma2014adam}) to optimize over both the hyperparameters $\alpha$ and parameters $w$; One could take a couple of gradient descent steps for $w$ and one step for $\alpha$ in an iterative fashion. There is no need to solve the inner optimization problem to completion. This will also result in significant speed-up by trading computation for more memory consumption. As a compromise between purely automated solutions to Eq. \ref{eq:multi_task} (i.e., AutoML) and the expert-in-the-loop solutions (i.e., MLOps), we could also use a combination of the two \cite{tan2019efficientnet, radosavovic2020designing}. Moreover, as an additional layer of complication, the performance metric $\mathcal{M}_{\text{val}}(w^*(\alpha), \alpha)$ could pursue multiple competing objectives such as minimizing the error rates (i.e., maximizing accuracy) and reducing the computational and memory costs of the model, perhaps to make them suitable for smaller devices such as mobile phones, tablets, or IoT (Internet of Things) devices \cite{tan2019mnasnet, howard2019searching}. Here, balancing the trade-off between optimizing one objective versus the other is an open problem. This will introduce additional hyper-parameters (lets call them ``hyper-hyper-parameters'') for the outer optimization problem in Eq. \ref{eq:multi_task}. Last but not least, data augmentation policies \cite{cubuk2019autoaugment, cubuk2020randaugment} could also be part of the search space (i.e., $\alpha$).\\

This should give us enough background on the current state-of-the-art in \href{https://www.youtube.com/playlist?list=PLoEMreTa9CNnQXiups8QMzmyKe4b3ge6F}{AutoML} and help us outline the open problems that need to be addressed collectively by the machine learning community.

\section{Open Problems}
As mentioned above, despite the clean and unified mathematical formulation of the iteration step as a bi-level optimization problem (see Eq. \ref{eq:multi_task} and Fig. \ref{fig:big_picture}), its solutions are case specific and complex. In the following, we will consider such cases while increasing the level of complexity from supervised learning to semi-supervised, self-supervised, unsupervised, few-shot, federated, reinforcement, and physics-informed learning. As a consequence of this exercise, this work exposes a multitude of open challenges in the field, many of which can be addressed in parallel.

\subsection{Supervised Learning}
Most of the progress made over the last few years, ever-since the advent of AutoML, falls under the umbrella of supervised learning, in particular (image) classification \cite{zoph2016neural, zoph2018learning, cubuk2019autoaugment, real2019regularized, miikkulainen2019evolving, pham2018efficient, liu2018darts, tan2019efficientnet, radosavovic2020designing, tan2019mnasnet, howard2019searching}. These works use accuracy as their performance metric or a differentiable approximation of it (e.g., the negative of the loss function or the log-likelihood). As a low hanging fruit, we could investigate if such techniques would still work in the face of other performance metrics, such as precision, recall, F1 score, calibration \cite{guo2017calibration}, etc., for balanced and unbalanced datasets. Finding differentiable approximations of such metrics is particlarly interesting because we can employ similar techniques to the ones used in \cite{liu2018darts} (i.e., using plain-vanilla stochastic gradient descent to optimize over both hyper-parameters and parameters of our models). Such techniques are interesting not only because of their computational efficiency but also because they require very little ``hyper-hyper-parameter'' tuning. However, how to make such methods memory efficient is still an open question. Moreover, methods such as Bayesian optimization, reinforcement learning, evolutionary algorithms, and even parameter-sharing for solving Eq. \ref{eq:multi_task} introduce additional hyper-parameters (i.e., ``hyper-hyper-parameter''); In simple terms, we would like to avoid doing AutoAutoML, AutoAutoAutoML, etc., and fragmenting our datasets beyond training, validation, and testing. We could also study the effect of such approximations to different performance metrics and provide theoretical upper-bounds on the loss of performance as a result of such approximations.\\

\textbf{\href{https://www.youtube.com/playlist?list=PLoEMreTa9CNlO37YBuFDU7_meLSjvSw2a}{Large Networks:}} Recently, we are witnessing a trend in computer vision trying to replace convolutions neural networks with transformers \cite{dosovitskiy2020image, touvron2021training, liu2021swin} or even multi-layer perceptrons \cite{tolstikhin2021mlp}, inspired by their success in language \cite{devlin2018bert, brown2020language, liu2021pay}. The question is if the currently available techniques (see the background section) for solving Eq. \ref{eq:multi_task} could generalize to such architectures and improve their performance. It is worth noting that the techniques outlined in the background section are primarily designed for convolutions neural networks. Furthermore, when it comes to data-augmentation strategies, the current techniques leverage only a single image \cite{cubuk2019autoaugment}, the question is if novel data-augmentation strategies such as mix-up \cite{zhang2017mixup} and cut-mix \cite{yun2019cutmix} (leveraging pairs of images) can be discovered as part of solving Eq. \ref{eq:multi_task}. Answering these two questions would entail rethinking the design space (i.e., the space in which $\alpha$ is assumed to live). Another interesting and fundamental question is investigating the possibility of automating the discovery of learning rate schedules such as the cosine learning rate schedule \cite{loshchilov2016sgdr}. Here, the learning rate is a function of the current epoch rather than being a constant. This will significantly increase the complexity of the iteration problem (see Eq. \ref{eq:multi_task}). Methods similar to the ones presented in \cite{liu2018darts} seem to have a good chance at solving this problem because they don't rely on solving the inner optimization problem (i.e., training) in Eq. \ref{eq:multi_task} to completion. This will allow us to modify the learning rate in tandem with the training process per each epoch.\\

\textbf{\href{https://www.youtube.com/playlist?list=PLoEMreTa9CNk2TXiWDl0i-Gsou3ejD7za}{Small Networks:}} There are times when not only we are looking for the most performant model but also we want the model to be as memory and compute efficient as possible. This is an important stepping stone towards democratizing artificial intelligence in anticipation of the future of Internet of Things where a lot of our devices (e.g., cellphones, cars, security cameras, refrigerators, air conditioners, etc.) will be intelligent. Such devices usually have smaller compute capabilities and memory capacity than our computers in data-centers or on the cloud. To make them intelligent we need to take their constraints into consideration. Mathematically speaking, $\mathcal{M}_{\text{val}}$ is a weighted combination of at least two objectives; One is the performance metric, while the other is about making the model more nimble and could take different forms such at FLOPs, MAC (memory access cost), number of parameters, latency and memory consumption of the target devices. This last item necessitates a hardware-in-the-loop approach. The weights given to each objective function are what we called ``hyper-hyper-parameters'' earlier in this document. It is still an open question how to set such weights in order to balance the tradeoff between optimizing one objective versus another. We are therefore dealing with a multi-objective bi-level optimization problem. Here, ideas such as the ones proposed in \cite{kendall2018multi} for multi-task learning using uncertainty to weigh different objectives could be extended to solve our multi-objective bi-level optimization problem. A similar multi-objective optimization problem arises in physics-informed deep learning \cite{raissi2019physics, Raissi2018DeepHP, Raissi2020HiddenFM} where we need to balance the trade-off between fitting the data and respecting the law of physics modeled using ordinary and partial differential equations. Alternatively, we could investigate the possibility of automating the process of making pre-trained models smaller. This is an after-the-fact approach where we would like to automate the discovery of methods such as knowledge distillation \cite{hinton2015distilling}, model pruning and compression \cite{han2015learning, han2015deep, rastegari2016xnor}, etc. This approach will avoid the aforementioned multi-objective bi-level optimization problem and will instead break the problem into two or more stages; In the first stage we will be looking for the most performant model, regardless of its cost, while in the second stage we will look for the best model compression strategies. This will lead to a multi-stage (versus multi-objective) bi-level optimization problem.\\

\textbf{\href{{https://www.youtube.com/playlist?list=PLoEMreTa9CNnBnw17OkCP70ASwmHQqvQt}}{Robustness:}} It is a well-known fact that deep neural networks are vulnerable to adversarial and backdoor attacks \cite{szegedy2013intriguing, goodfellow2014explaining, kurakin2018adversarial, su2019one, bagdasaryan2020backdoor}. We could be investigating how much of this vulnerability can be attributed to the iteration step (see Fig. \ref{fig:big_picture} and Eq. \ref{eq:multi_task}) and whether automating the process can alleviate or aggravate the problem.\\

\textbf{\href{https://www.youtube.com/playlist?list=PLoEMreTa9CNkOG-DEF1RQ_oPXKGTO988o}{Explainable AI:}} In the past few years, the machine learning community has made a lot of progress in the emerging field of explainable and trustworthy AI (see e.g., \cite{zeiler2014visualizing, simonyan2014deep, springenberg2014striving, montavon2018methods, ribeiro2016should, zhou2016learning, zhang2021understanding, selvaraju2017grad, lundberg2017unified, shrikumar2017learning, sundararajan2017axiomatic, guo2017calibration, bau2020understanding, raghu2021vision} and the references therein) to the extend that deep neural networks are no longer considered black boxes but rather gray ones. However, there has been very little effort in the literature (if any) to explain the choices made as part of the iteration process (see Fig. \ref{fig:big_picture}). The question is what features of the data, or rather the meta-data, explain the choices we make for learning rate schedules, architecture designs, data-augmentation, etc. What is the effect of noise in the data on such choices? How important are the size and intrinsic dimensionality of the data? This a place where the underlying structure of the data (i.e., the lower dimensional manifold on which the data lives) could help us shed some light on these fundamental questions. If successful, efforts in this direction could lead to a new field, namely ``Explainable AutoML''.\\

\textbf{\href{https://www.youtube.com/playlist?list=PLoEMreTa9CNkn3ofeGWkuLllzdbVPMqWS}{Transfer Learning:}} Moreover, in a parallel thrust, we could be studying the impact of the iteration process (see Fig. \ref{fig:big_picture}) on the generalizability and transferability of the learned features to downstream tasks. The field of deep learning is largely driven by the ideas of Transfer learning to the extent that we rarely train our models from scratch. Along the same lines, an important question worthy of systematic studies is transferability (or lack thereof) of data augmentation policies (e.g., AutoAugment and RandAugment) from one dataset (e.g., ImageNet) to another (e.g., Pascal VOC).\\

\textbf{\href{https://www.youtube.com/playlist?list=PLoEMreTa9CNkh9JAoKa9TwO2LufHYdQSz}{Semantic Segmentation:}} When it comes to the task of semantic segmentation, we typically start with neural networks pre-trained on a related classification task and do transfer learning \cite{long2015fully, noh2015learning, ronneberger2015u, chen2017deeplab, yu2015multi, lin2017refinenet, chen2018encoder, fu2019dual, zheng2021rethinking}. This is mainly because we usually have access to smaller datasets for this task as labeling every single pixel in an image is more cumbersome than labeling an entire image with a single label. Referring back to Fig. \ref{fig:big_picture}, there is very little work on an iteration phase dedicated to the semantic segmentation task. Here, the performance metrics (e.g., pixel accuracy, mean accuracy, mean IU, frequency weighted IU, etc.) are more complex than the accuracy metric often used for the classification task. Finding differentiable approximations of such metrics so that we can employ similar techniques to the ones used in \cite{liu2018darts} is particularly interesting. Moreover, when it comes to the task of semantic segmentation, we need to not only capture the global information in an image (i.e., resolving the ``what'' of the image) but also the local information (i.e., resolving the ``where'' of the objects in the image). It would be interesting to investigate how the iteration stage (see Fig. \ref{fig:big_picture}) would have an impact on the ``what'' and ``where'' components of the semantic segmentation task. In particular, would the iteration phase leverage tools such as atrous convolutions, short-cut connections, conditional random fields, multi-scale aggregation, deep supervision, deconvolutions, upsampling, attention mechanisms, etc.? If so, to what extent?\\

\textbf{\href{https://www.youtube.com/playlist?list=PLoEMreTa9CNl6RQNFrHnxNWzq3Byy8bAd}{Super-Resolution, Denoising, and Colorization:}} When it comes to creative tasks such as super-resolution, denoising, colorization, and style transfer where the output of a neural network is an image, it is very hard to judge the quality of the generated images in a quantitative fashion. Available perormance metrics such as Peak Signal-to-Noise (PSNR), Structural Similarity (SSIM), and Feature Similarity (FSIM) fall short of doing justice to the task. It is therefore very hard to measure progress in these fields \cite{dong2015image, johnson2016perceptual, gatys2016image, huang2017arbitrary, kim2016accurate, shi2016real, zhang2017beyond, lim2017enhanced, ulyanov2018deep, zhang2018residual, zhang2018image, zhang2018unreasonable, zhang2016colorful} and more importantly guide the iteration phase (see Fig. \ref{fig:big_picture}). In this regard, deep features extracted from deep neural networks (e.g., VGG) trained on the ImageNet classification task show unreasonable empirical effectiveness as perceptual metrics \cite{zhang2018unreasonable}. However, there aren't many works (if any) that guide the iteration phase using such deep features (e.g., Learned Perceptual Image Patch Similarity (LPIPS) metric) in an automated fashion. In a related note, the tasks of super-resolution, denoising, colorization, and style transfer usually entail balancing the trade-off between multiple loss functions (e.g., reconstruction L1/L2 loss versus perceptual loss \cite{johnson2016perceptual}). It is still an open problem how to strike such a balance in the absence of universally accepted performance metrics. Here, ideas such as the ones proposed in \cite{kendall2018multi} for multi-task learning using uncertainty to weigh different objectives could be extended to solve this problem. Alternatively, we could investigate the possibility of guiding the iteration process (see Fig. \ref{fig:big_picture}) using human-in-the-loop Reinforcement Learning algorithms where the reward signal comes from the judgment of human beings; A human can easily take a look at an image and associate a quality score to it, perhaps from 1-10. This is feasible because with Reinforcement Learning we don't need to differentiate through the reward signals or the thought process of the human evaluator. Human-in-the-loop techniques are gaining traction these days (see e.g., \cite{wang2022skill}) because writing well-defined reward functions is very challenging if not impossible for many real-word applications of Reinforcement Learning beyond games and simulated environments.\\

\textbf{\href{https://www.youtube.com/playlist?list=PLoEMreTa9CNmPGaVQYDWydc2ZmEt6zdrV}{Pose Estimation:}} A keyboard and a mouse are not the only means of interacting with a computer; A key topic in the field of human-computer interaction in particular and the meta verse in general is human pose estimation \cite{toshev2014deeppose, wei2016convolutional, newell2016stacked, cao2017realtime, sun2019deep}. Two evaluation metrics that could guide the iteration phase (see Fig. \ref{fig:big_picture}) are Percentage of Correct Parts (PCP) and Percent of Detected Joints (PDJ). PCP measures detection rate of limbs, where a limb is considered detected if the distance between the two predicted joint locations and the true limb joint locations is at most half of the limb length. As for PDJ, a joint is considered detected if the distance between the predicted vector and the true joint is within a certain fraction of the torso (e.g., left shoulder and right hip) diameter. A closely related metric is Percentage of Correct Keypoints (PCK) which measures the percentage of detections that fall within a normalized distance of the ground truth. It would be interesting to investigate how the iteration stage (see Fig. \ref{fig:big_picture}) would leverage these metrics or their differentiable approximations to come up with novel architectures (e.g., stacked hourglass blocks, cascaded pyramid layers, part affinity fields, etc.) in an automated fashion. This is not a well-studied topic as of this writing.\\

\textbf{\href{https://www.youtube.com/playlist?list=PLoEMreTa9CNnHJYvv5djNWvDR0rtec_Vb}{Optical Flow and Depth Estimation:}} The world around us is 3D and evolving in time. On the one hand, depth estimation \cite{eigen2014depth, eigen2015predicting, godard2017unsupervised, zhou2017unsupervised, kopf2021robust} allows us to add a third dimension to our 2D images and has applications for self-driving cars and robots. On the other hand, optical flows \cite{fischer2015flownet, ilg2017flownet, sun2018pwc} enable us to capture the evolution in time and the relationship between consecutive frames in a video. In addition to applications for self-driving vehicles and robotics, optical flows can be used as additional features for the task of action recognition in videos. For optical flows, End Point Error (EPE) is typically used as the performance metric guiding the iteration phase (see Fig. \ref{fig:big_picture}). It is the Euclidean distance between the predicted flow vector and the ground truth, averaged over all pixels. Here, the training data is usually simulated because it is very hard to measure optical flows in the physical world \cite{fischer2015flownet}. Inevitably, we have to rely on domain adaptation techniques to close the reality gap (i.e., the gap between the real and simulated data distributions) to the extent possible. It is therefore an intriguing research question to study the effect of domain adaptation techniques on the iteration phase (see Fig. \ref{fig:big_picture}) and vice versa. The task of depth estimation also suffers from lack of enough labeled training data. Fortunately, there are ways to leverage the underlying physics of the problem to perform unsupervised monocular depth estimation with e.g., left-right consistency \cite{godard2017unsupervised}. Here, the training loss would involve a weighted combination of multiple individual loss functions \cite{godard2017unsupervised, zhou2017unsupervised}. Properly setting those weights in an automated fashion is an open question. Here, we could utilize performance metrics such as absolute relative distance, squared relative distance, or root mean square error (RMSE) to guide the iteration stage (see Fig. \ref{fig:big_picture}).\\

\textbf{Object Detection:} Generally speaking, there are two major types of object detectors: \href{https://www.youtube.com/playlist?list=PLoEMreTa9CNm18TPHIYm3t2CLIqxLxzYD}{multi-stage} (typically two) and \href{https://www.youtube.com/playlist?list=PLoEMreTa9CNlDc-55lDiKpBb5wUZ_oHqO}{one-stage} detectors. With multi-stage detectors \cite{padilla2020survey, girshick2014rich, he2015spatial, girshick2015fast, ren2015faster, dai2016r, lin2017feature, dai2017deformable, he2017mask, cai2018cascade}, the primary objective is to find the most performant model (measured using mean average precision) while efficiency (measured using frames per second) is a secondary objective. With single-stage detectors \cite{sermanet2013overfeat, redmon2016you, liu2016ssd, redmon2017yolo9000, lin2017focal, huang2017speed, redmon2018yolov3, law2018cornernet, tian2019fcos, zhou2019objects, tan2020efficientdet, bochkovskiy2020yolov4, carion2020end, zhu2020deformable}, the primary objective is to find the most agile model (measured using frames per second) while performance (measured using mean average precision) is a secondary objective. Referring back to equation \ref{eq:multi_task}, $\mathcal{M}_{\text{val}}$ is pursuing multiple (i.e., at least two) competing objectives for the task of object detection. Balancing the trade-off between optimizing one objective versus the other is an open problem. It is worth mentioning that this problem is about what we called ``hyper-hyper-parameters'' in the iteration phase and that we would like to avoid fragmenting our datasets beyond training, validation and testing. As for hyper-parameters $\alpha$, the design space (the space in which $\alpha$ lives) is a much more complex one for object detection compared to classification tasks. The input data could be in the form of images, image patches, image pyramids, etc. The backbone could be in the form of VGG, ResNet, ResNeXt, Darknet, Hourglass Network, Transformers, etc. The neck of the architecture could be in the form of FPN (Feature Pyramid Network), PANet, Bi-FPN (Bi-directional FPN), etc. The head of the object detection system could be a dense predictor (RPN, YOLO, SSD, RetinaNet, FCOS) or a sparse one (Faster R-CNN, R-FCN). As for data augmentation we could use CutMix, MixUp, Mosaic, Bluring, etc. For the loss functions, we could use L1, L2, Smooth L1, or CIoU for the regression component of the total loss function and MSE, binary or multi-class cross-entropy loss for the classification portion of the total loss. Therefore, exploring the design space in a systematic and automatic fashion as part of the iteration phase (see Fig. \ref{fig:big_picture}) is a challenging task. Last but not least, all metrics are wrong, some are useful. This is specially true when it comes to the object detection task. There are heated debates in the literature about the appropriateness of mean average precision (mAP) or its COCO style variants as valid performance metrics for the object detection task (see e.g., \cite{redmon2018yolov3}).\\

\textbf{\href{https://www.youtube.com/playlist?list=PLoEMreTa9CNnnBrqy4cJSZ3NX3rhvcJxi}{Face Recognition and Detection:}} Face detection is a special case of the object detection and key point (pose) estimation topics that we covered earlier in this document. Futhermore, face recognition (verification and identification) can be viewed as a close-set or an open-set problem, depending on the type of available data. Close-set face recognition is nothing but a classification task that we covered earlier in this document. Open-set face recognition on the other hand is about metric learning, i.e., learning features that are capable of pulling similar images together while pushing dissimilar images apart. The literature on open-set face recognition spends a lot of time designing new loss functions such as the triplet loss \cite{schroff2015facenet, hermans2017defense}, the center loss \cite{wen2016discriminative}, angular softmax loss \cite{liu2017sphereface}, additive angular margin loss \cite{deng2019arcface}, etc. It is therefore a natural question to ask if it is possible to automate the search for appropriate loss functions as part of the iteration phase (see Fig. \ref{fig:big_picture}). To guide the iteration process we could leverage the ROC (Receiver Operator Characteristic) curves relating the true positive rate to the false positive rate.\\

\textbf{\href{https://www.youtube.com/playlist?list=PLoEMreTa9CNkvCOt7-WCh0xEBICaqHo6Y}{Video} \& \href{https://www.youtube.com/playlist?list=PLoEMreTa9CNmvjhZTBWfyMuFggNKkIYKa}{3D data:}} As mentioned earlier in this document, the world around us is evolving in time and is 3D. When it comes to videos, we could think of at least two important applications, namely action recognition and object tracking. Action recognition \cite{ji20123d, karpathy2014large, simonyan2014two, tran2015learning, wang2015action, wang2016temporal, feichtenhofer2016convolutional, Carreira_2017_CVPR, Tran_2018_CVPR, Feichtenhofer_2019_ICCV} is a classification task albeit on a sequence of image frames in a video as the input data. Here the design space (the space in which $\alpha$ lives, referring to equation \ref{eq:multi_task}) is more complex compared to the design space for images. It would therefore be interesting to see if ideas such as early/late/slow fusion, multi-streaming, using optical flows as additional input features, 3D convolutions, trajectory pooling, or slow-fast networks would survive the iteration phase (see Fig. \ref{fig:big_picture}). Furthermore, when it comes to object tracking \cite{nam2016learning, bertinetto2016fully}, there is very little work on automating the iteration phase (see Fig. \ref{fig:big_picture}) and studying its impact on the resulting algorithms. Here, we could use evaluation metrics such as the center location error and the bounding box overlap ratio to guide the iteration phase. In a parallel thrust, performing object recognition, detection and segmentation on 3D point cloud data is more challenging than doing so on images \cite{Qi_2017_CVPR, qi2017pointnetplusplus, Wang2019DynamicGC, Zhao2021PointT, Zhou_2018_CVPR}. These types of data (e.g., LIDAR data) appear naturally in self-driving vehicles and robotics applications. However, there is very little work on automating the iteration phase (see Fig. \ref{fig:big_picture}) for performing object recognition, detection and segmentation on 3D point cloud data.

\subsection{Beyond Supervised Learning}
It is now time to increase the level of complexity and move beyond supervised learning towards semi-supervised, self-supervised, unsupervised, few-shot, federated, reinforcement, and physics-informed learning. We will be approaching these topics from the perspective of the iteration phase (see Fig. \ref{fig:big_picture}).\\

\textbf{Natural Language Processing:} For applications such as \href{https://www.youtube.com/playlist?list=PLoEMreTa9CNmUtcQuVlcAMMpGxkJG3rp-}{word vector representations} \cite{Mikolov2013LinguisticRI, Mikolov2013DistributedRO, Mikolov2013EfficientEO,Pennington2014GloVeGV, Bojanowski2017EnrichingWV}, \href{{https://www.youtube.com/playlist?list=PLoEMreTa9CNnfXd5mAsQlaD-NGQS0r4_2}}{text classification and sequence tagging} \cite{Socher2013RecursiveDM, Kim2014ConvolutionalNN, Le2014DistributedRO, Johnson2015EffectiveUO, Kalchbrenner2014ACN, Zhang2017ASA, Zhang2015CharacterlevelCN, Tai2015ImprovedSR, Joulin2017BagOT, Yang2016HierarchicalAN, Huang2015BidirectionalLM, Lample2016NeuralAF, Ma2016EndtoendSL, Howard2018UniversalLM}, \href{https://www.youtube.com/playlist?list=PLoEMreTa9CNlYq4uIDkbluDVa8_y2obh_}{translation} \cite{Bahdanau2015NeuralMT, Sutskever2014SequenceTS, Cho2014LearningPR, Cho2014OnTP, Luong2015EffectiveAT, Sennrich2016NeuralMT, Wu2016GooglesNM, Gehring2017ConvolutionalST, Vaswani2017AttentionIA, Kudo2018SentencePieceAS, Kudo2018SubwordRI, Kitaev2020ReformerTE, Katharopoulos2020TransformersAR, Choromanski2021RethinkingAW} and \href{https://www.youtube.com/playlist?list=PLoEMreTa9CNkVn8pKuRu-Ql1zAgVVv3kh}{language modeling} \cite{Peters2018DeepCW, Bai2018AnEE, Radford2018ImprovingLU, Devlin2019BERTPO, Radford2019LanguageMA, Lan2020ALBERTAL, Liu2019RoBERTaAR, Sanh2019DistilBERTAD, Dai2019TransformerXLAL, Yang2019XLNetGA, Raffel2020ExploringTL, Gururangan2020DontSP, Lample2019CrosslingualLM, Conneau2020UnsupervisedCR, Joshi2020SpanBERTIP, Lewis2020BARTDS, Beltagy2020LongformerTL, Reimers2019SentenceBERTSE, Brown2020LanguageMA, Clark2020ELECTRAPT, Gao2021SimCSESC, Liu2021PayAT, Chen2021EvaluatingLL, Holtzman2020TheCC}, unlabeled text data is available en masse thanks to the Internet -- for example, the \href{http://commoncrawl.org}{Common Crawl} project produces about 20TB of text data extracted from web pages each month. This makes pre-training large language models on such data particularly attractive. Such models can then be fine-tuned on downstream tasks (BERT family of models) or used in a few-shot setting (GPT type of models). Here, the fundamental question is what performance metric(s) should we use to guide the iteration phase (see Fig. \ref{fig:big_picture}) when it comes to training large language models on unlabeled data. One idea is to use perplexity as a measure of the goodness of language models. However, we don't typically train large language models for the sake of modeling the language but rather to use them in some downstream applications such as text classification (e.g., sentiment analysis), sequence tagging (e.g., named entity recognition), machine translation, program synthesis, question answering, summarization, semantic textual similarity, language comprehension, conversational response generation, etc. For instance, it is a well-known observation that the common practice of extracting sentence embeddings from the BERT language model, by average pooling the last layer output vectors or using the output of the first token (i.e., the [CLS] token), yields rather bad sentence embeddings \cite{Reimers2019SentenceBERTSE}, often worse than averaging the GloVe vectors. This is why researchers came up with the ideas of Sentence-BERT and Siamese BERT-Networks. Another observation is that BERT in its original form cannot perform translation. This is why researchers introduced BART, GPT, T5, etc. However, each one of these contributions focus on a hand-full of downstream tasks to evaluate the performance of their language modeling capabilities. Some focus on the GLUE benchmark \cite{Wang2018GLUEAM} of a suit of downstream tasks, some focus on the BLUE score \cite{Papineni2002BleuAM} for translation, etc. Perhaps a better strategy to guide the iteration process (see Fig. \ref{fig:big_picture}) is to approach language modeling from a multi-task learning perspective where $\mathcal{M}_{\text{val}}$ in equation \ref{eq:multi_task} is a weighted combination of the performance metrics for a multitude of downstream tasks. Here, an open question is how to properly weigh one objective function versus the others. Here, ideas such as the ones proposed in \cite{kendall2018multi} for multi-task learning using uncertainty to weigh different objectives could be extended to solve our multi-objective bi-level optimization problem. In addition, we could include extra objectives in $\mathcal{M}_{\text{val}}$ to penalize the computational complexity and memory consumption of the resulting language models. The idea is to come up with the smallest language model that is good at solving a multitude of downstream tasks in an automated fashion. Of particular interest are techniques similar to the ones used in \cite{liu2018darts} (i.e., using plain-vanilla stochastic gradient descent to optimize over both hyper-parameters and parameters of our models). Last but not least, many of the ideas in natural language processing can be extend to \href{https://www.youtube.com/playlist?list=PLoEMreTa9CNnbzENK14asgbwIGmNGvdC_}{graphs} (e.g., social networks) \cite{Bordes2013TranslatingEF, Perozzi2014DeepWalkOL, Tang2015LINELI, Grover2016node2vecSF, Kipf2017SemiSupervisedCW, Defferrard2016ConvolutionalNN, Hamilton2017InductiveRL, Velickovic2018GraphAN, Xu2019HowPA, Schlichtkrull2018ModelingRD}.\\

\textbf{\href{https://www.youtube.com/playlist?list=PLoEMreTa9CNkCnmkOa9Av0KgEQmVWhMkp}{Multimodal Learning:}} With multimodal learning \cite{Donahue2015LongtermRC, Vinyals2015ShowAT, Karpathy2017DeepVA, Xu2015ShowAA, Ba2016LayerN, Anderson2018BottomUpAT, Reed2016GenerativeAT, Zhang2017StackGANTT, Radford2021LearningTV, Ramesh2021ZeroShotTG, Jaegle2021PerceiverIA, ramesh2022hierarchical}, we are taking baby steps towards human level artificial intelligence (i.e., artificial general intelligence). For instance, if we look at an intelligent robot and say ``pick that up and put it on the table'' while pointing at a box sitting on the ground, to be able to execute the command correctly, the robot should not only process speech and language but also should be able to use its vision system to understand what we mean when we say ``that''. Moreover, a common criticism to large language models is that even if they manage to generate seemingly cohesive text, they have very little idea about what they are actually talking about; For example, a language model trained only on textual data has never seen images of airplanes, cars and ships, it has only read about them on the internet. The field of multimodal learning is therefore attracting the attention of a lot of great researchers both in academia and the industry. Two important applications are translating images (or videos) to text (e.g., image and video captioning) and vice versa (e.g., text to image synthesis). Another equally important application is visual question answering. Moreover, training large language models both on images and textual data is also showing some great promise. We will be approaching these topics from the perspective of the iteration phase (see Fig. \ref{fig:big_picture}). When it comes to translating images to text, we could use the BLUE score to guide the iteration phase. However, there are heated debates in the literature if the BLUE score is the best performance metric both for image captioning and translation. Coming up with better metrics is an open problem. Moreover, there is very little work (if any) on automating the iteration phase of translating images to texts. For the text to image synthesis type of tasks, like any other creative task (e.g., super-resolution, denoising, colorization, style transfer and generative adversarial networks), it is very hard to judge the quality of the generated images in a quantitative fashion. Here, we could use metrics such the Inception Score, the Fréchet Inception Distance or the Learned Perceptual Image Patch Similarity (LPIPS) metric to guide the iteration phase. As of this writing, there aren't many works (if any) that have done so before. Furthermore, when it comes to training large language models on unlabeled textual data as well as images, the fundamental question is what performance metric(s) should we use to guide the iteration phase (see Fig. \ref{fig:big_picture}). In particular, the question is how existing performance metrics such as perplexity can be generalized to handle both text and image type of data. Alternatively, similar to language models trained only on text, we could define a set of downstream tasks involving both text and images as benchmarks to guide the iteration phase. The idea is then to approach multimodal modeling from a multi-task learning perspective.\\

\textbf{Generative Networks:} Generative models are either trained to maximize the likelihood (or rather its lower bound) of the training data (e.g., \href{https://www.youtube.com/playlist?list=PLoEMreTa9CNlgs3IByUnTIp6T4uk9vlax}{Variational Auto-Encoders}) or minimize a distance/divergence between the training data distribution and model predictions (e.g., \href{https://www.youtube.com/playlist?list=PLoEMreTa9CNnYMbO5DaypRM-AU_nvJgT2}{unconditional} and \href{https://www.youtube.com/playlist?list=PLoEMreTa9CNlIopdozBBp04xPNJspaaqC}{conditional} Generative Adversarial Networks). A central question here is how to measure the quality of the generated data in a quantitative fashion. For images, we could use the Inception score or the Fréchet Inception Distance to guide the iteration process, despite all their imperfections. However, it is not clear what performance metrics we should use for other types of data such as text, speech, graphs (e.g., social networks), etc. Furthermore, it is a well-known observation that training generative adversarial networks is an unstable process and most of the contributions in this field are made towards stabilizing this process by using different loss functions (e.g., Feature Matching, Least Squares GANs, Wasserstein GANs, Hinge Loss, etc.), normalization and regularization schemes (e.g., Gradient Penalty, Spectral Normalization, Orthogonal Regularization, Adaptive Instance Normalization, Path Length Regularization, etc.), architectures (e.g., DCGANs, Self-Attention GANs, etc.) and training schedules (Progressive Growing, two time-scale update rule, historical averaging of parameters, etc.). It would be interesting to study if such techniques would survive an automated iteration process (see Fig. \ref{fig:big_picture}). Of particular interest is the progressive growing idea because the neural network architecture itself is a function of the current training epoch. This will significantly increase the complexity of the iteration problem (see Eq. \ref{eq:multi_task}). Methods similar to the ones presented in \cite{liu2018darts} seem to have a good chance at solving this problem because they don't rely on solving the inner optimization problem (i.e., training) in Eq. \ref{eq:multi_task} to completion. This will allow us to modify the architecture in tandem with the training process per each epoch. Last but not least, when it comes to conditional GANs (e.g., image-to-image translation), we typically try to minimize a total loss function being a weighted combination of multiple individual loss functions. It is still an open problem how to come up with those weights in the absence of validation data and appropriate performance metrics. Perhaps reformulating the problems as a multi-task learning problem and using ideas similar to the ones proposed in \cite{kendall2018multi}, that leverage uncertainty to weigh different objectives, could help us address this problem.\\

\textbf{\href{https://www.youtube.com/playlist?list=PLoEMreTa9CNmk1fcAYTQ1ps-GPkFmKA1-}{Domain Adaptation:}} The concept of domain adaptation \cite{long2015learning, ganin2016domain, tzeng2017adversarial, bousmalis2017unsupervised, hoffman2018cycada} is related to scenarios where we have a lot of labeled data (e.g., simulated data) from a source domain and zero (or very few) labeled data (e.g., real data) from a target domain. Such scenarios happen frequently in many engineering fields (sometimes called multi-fidelity modeling \cite{Perdikaris2017NonlinearIF, Raissi2016DeepMG, raissi2014multi} in fluid and solid mechanics) including but not limited to self-driving cars and robotics. Domain adaptation can help us close the so called reality gap between the simulated and real data distributions. We are going to approach domain adaptation from the perspective of the iteration phase (see Fig. \ref{fig:big_picture}). With domain adaptation we would like to minimize the risk of making errors on the target data, not necessarily the source data. Now, the question is how can we measure the performance of our models on the target data in the absence of any target labels (e.g., unsupervised domain adaptation) or in the presence of very few of them (e.g., weakly-supervised domain adaptation). One idea is to use unsupervised hyper-parameter selection techniques \cite{ganin2016domain}. We can split the source domain labeled data into a training set $S_\text{train}$ and a validation set $S_\text{val}$. Similarly, we can split the target domain unlabeled data into a training set $T_\text{train}$ and a validation set $T_\text{val}$. We can then use $S_\text{train}$ and $T_\text{train}$ to learn a model using domain adaptation techniques. The trained model can now be used to generated labels for the unlabeled data $T_\text{train}$. We then remove the labels from $S_\text{train}$. We will then train a reverse model using $T_\text{train}$ as the source domain and $S_\text{train}$ as the target domain. The reverse model can now be evaluated on the validation set $S_\text{val}$. We could use this reserve validation risk as a proxy for the true validation risk to guide the iteration process.\\

\textbf{\href{https://www.youtube.com/playlist?list=PLoEMreTa9CNkw8PX3mgEyXLLLOKZ5j-ia}{Few-shot Learning:}} When it comes to using a machine learning model to serve millions of users perhaps over the internet, not only we need to take care of the distributional shift between the training and test data (i.e., domain adaptation) but also we need to be able to handle new use cases and more importantly new labels associated to such use cases. As an example, in the context of a \href{https://www.youtube.com/playlist?list=PLoEMreTa9CNkeu48fSTkHScJaTfn46C_7}{recommendation system} \cite{Hidasi2016SessionbasedRW, Sedhain2015AutoRecAM, Cheng2016WideD, He2017NeuralCF, He2017NeuralFM, Guo2017DeepFMAF, Quadrana2017PersonalizingSR, Liang2018VariationalAF, Tang2018PersonalizedTS, Naumov2019DeepLR}, we can think of new items (e.g., movies, products, etc.) to be recommended to new/existing users. This is related to the topic of few-shot learning \cite{vinyals2016matching, snell2017prototypical, sung2018learning} where our models need to be able to handle new labels given very few observations per each label. For classification tasks, we can use $N$-way $K$-shot classification accuracy to guide the iteration phase (see Fig. \ref{fig:big_picture}). It would therefore be interesting to study the effect of an automated iteration stage on the resulting algorithms. More importantly, it is still unclear what performance metrics we should use for other applications of few-shot learning beyond classification such as semantic segmentation, object detection, pose estimation, depth estimation, etc.\\

\textbf{\href{https://www.youtube.com/playlist?list=PLoEMreTa9CNmPnHffTFkPpbCJWpzpm4ln}{Federated Learning:}} In addition to the publicly available data on the internet, there is a wealth of data sitting on our privately-held devices (e.g., cellphones, tablets, laptops, etc.). Our devices are also getting more powerful in their compute and data collection capabilities (e.g., multiple camera lenses on the back of our cellphones). Federated Learning tries to leverage such privately held data to train machine learning models while preserving the privacy of the data. The idea is to bring the models to the data (rather than bringing the data to the models on the cloud) and use the heterogeneous compute capabilities of user devices to train our models. What is being communicated to the cloud is the parameters of our models or rather their gradients. The field is still in its infancy and there are many open technical and non-technical challenges (e.g., communication efficiency, the non-i.i.d nature of the data, data pre-processing, training self-supervised models, privacy preserving, being robust to backdoor attacks, being able to train models on smaller devices, etc.) to be addressed before we can fully realize the potential of federated learning. In this work, we are approaching federated learning from the iteration perspective (see Fig. \ref{fig:big_picture}). Given the distributed nature of the data over millions of user devices, the question is how can we evaluate the performance of our models. One idea is to have the users of our models give star ratings (perhaps out of five) to our models. We can then aggregate these stars as a feedback signal to guide the iteration phase. Given that such a performance metric is discrete and non-differentiable we can use methods based on Reinforcement Learning to perform hyper-parameter selection.\\

\textbf{\href{https://www.youtube.com/playlist?list=PLoEMreTa9CNn_t0-6tffb2ocSpsZOit8Y}{Semi-Supervised} \& \href{https://www.youtube.com/playlist?list=PLoEMreTa9CNnNn2vsAmBtWML6dVlIwbLI}{Self-Supervised} Learning:} Let us now move towards the cases where we have access to a lot of unlabeled data and very few labeled data, if any. We faced a similar situation in natural language processing. However, with language we are working with discrete tokens which makes it easier to perform self-supervision by defining next token prediction (e.g., GPT style models) or masked token prediction tasks (e.g., BERT style models). Discrete tokens make it possible to use the softmax function as the last layer of a neural network (e.g., a Transformer architecture) and turn the self-supervision problem into a classification one. However, image and speech type of data are continuous signals. Normal, Laplace, or even mixture of Gaussians for modeling the distribution of continuous random variables are not as flexible as the softmax function is in modeling the distribution of discrete random variables. It is therefore required to rethink the semi-supervised and self-supervised learning paradigms when it comes to continuous signals (e.g., images and speech). In fact, this field has been growing at an exponential rate over the past two or three years. One common theme emerging in the literature is to take a single image and augment it into two different views. These two views should then give consistent representations once processed by the same neural network (or two similar ones). Here, a central challenge that needs to be overcome is avoiding the trivial solutions, either implicitly or explicitly. A network being supervised by itself or by another similar network is prone to converging to a trivial solution (e.g., a constant function ignoring its inputs altogether). We face a similar problem with physics-informed neural networks (e.g., any constant function is a solution to the Navier-Stokes equations) \cite{raissi2019physics, Raissi2018DeepHP, Raissi2020HiddenFM}. We are going to approach this problem from the perspective of the iteration stage (see Fig. \ref{fig:big_picture}). We need to either explicitly include a term in our training objective function that encourages non-trivial solutions, or design our search space in such a way that it includes mechanisms that have shown empirical success in avoiding trivial solutions such as stop-gradients, predictor heads, model averaging, contrastive losses, etc. Here, balancing the trade-off between the consistency loss and avoiding the trivial solution is very delicate. Fortunately, with semi-supervised learning \cite{miyato2018virtual, tarvainen2017mean, berthelot2019mixmatch, xie2020self, sohn2020fixmatch, touvron2021training} we have some labeled data and performance metrics that we can leverage to guide the iteration phase. However, with self-supervised learning \cite{caron2018deep, henaff2020data, tian2020contrastive, he2020momentum, misra2020self, chen2020simple, khosla2020supervised, chen2020big, grill2020bootstrap, caron2020unsupervised, chen2021exploring, bao2021beit, bardes2021vicreg, li2022dit, hamilton2022unsupervised}, neither such labeled data exists nor are there any universally accepted performance metrics. One idea is to define a set of downstream tasks (e.g., object recognition, detection, and segmentation for images) to judge the transferablility of the learned features. Here, how much importance we should give to each downstream task is an open problem.\\

\textbf{\href{https://www.youtube.com/playlist?list=PLoEMreTa9CNkL5fCXx50MN6nwWvsJgio6}{Speech:}} Similar to the language modeling paradigm for text, there is an emerging trend over the past few years to model speech \cite{oord2018representation, baevski2020wav2vec, hsu2021hubert, baevski2022data2vec, nguyen2022generative}. The idea is that before we (as human beings) learn to read and write, we learn to listen and speak. This self-supervised learning paradigm for speech is sometimes also called learning by listening. Speech being a continuous signal inherits many of the challenges that we went over in the previous paragraph on self-supervised learning for images. In addition, as of this writing, there are only two well-defined downstream tasks, namely translating speech to text \cite{Graves2006ConnectionistTC, graves2013speech, Graves2014TowardsES, hannun2014deep, greff2016lstm, amodei2016deep, Snyder2018XVectorsRD, park2019specaugment, li2019jasper} and vice versa \cite{graves2013generating, chung2014empirical, oord2016wavenet, kong2020hifi}, to guide the iteration stage (see Fig. \ref{fig:big_picture}) using their respective performance metrics (i.e., label error rate for speech recognition and subjective 5-scale mean opinion score in naturalness for speech synthesis). The performance metric for speech synthesis, however, requires human in the loop evaluators and leaves Reinforcement Learning or evolutionary algorithms as the only options to guide the iteration phase. Fortunately, there are some researchers in both academia and industry who are trying to come up with more downstream tasks to judge the quality of speech models. Even in the presence of such downstream tasks, we will need to solve multi-objective bi-level optimisation problems of a form that generalizes the one given in equation \ref{eq:multi_task}. This is still an open problem.\\

\textbf{\href{https://www.youtube.com/playlist?list=PLoEMreTa9CNl3yOJUx5oCmhCG_7Xu1Uml}{Reinforcement Learning:}} If we take a closer at the literature, many of the success stories of Reinforcement Learning are for Games (e.g., Atari, Chess, Shogi, Go, and StarCraft II) or in simulated environments (e.g., \href{https://github.com/openai/gym}{OpenAI Gym}, \href{https://mujoco.org}{MuJoCo}, etc.). For such cases, we have well-defined reward functions and are able to interact with the environment as many times as we like to collect enough experiences (i.e., data). This is not the case in the real world due the well-known physical constraints. What makes Reinforcement Learning difficult is 1) the need to collect plenty of experiences (i.e., inefficient use of data), 2) lack of well-defined rewards signals to not only guide the training process but also the iteration process (see Fig. \ref{fig:big_picture}) and 3) the sheer number hyper-parameters (i.e., degrees of freedom). We will therefore approach Reinforcement Learning from the perspective of the iteration process (see Fig. \ref{fig:big_picture}). A central question here whose answer can address (at least partially) all three of the aforementioned challenges is how to properly balance the trade-off between exploration and exploitation in the absence of well-denied reward functions. Here, we will investigate the possibility of using human-in-the-loop reward signals \cite{wang2022skill} (see also \href{https://openai.com/blog/chatgpt/}{ChatGPT}); A human can easily take a look at the performance of a robot in the real world and give it feedback. It is worth noting that this is different from imitation learning. To deal with the data-inefficiency issue, we can reformulate Reinforcement Learning as a multi-task learning problem; One task is to make the human evaluator happy and the other is to encourage exploration. Balancing the weights given to each objective is an open problem and has a direct impact on becoming more data-efficient.\\

\textbf{\href{https://www.youtube.com/watch?v=GXZq2_WYRjo}{Physics-Informed Learning:}} So far, we have been working on the brain of our artificial intelligent agents. If we take this brain, mount it on a robot (e.g., a drone) and ask it to operate in the real physical world (e.g., in a fluid), it will most definitely fail. This is because it has never learned to respect the laws of physics (e.g., conservation of mass, momentum and energy, gravity, etc.). If anything, it has learned to find loopholes of the simulated environment and bypass such laws (e.g., go faster than the speed of light), simply because it is trained only to maximize a reward signal or fit the corresponding data. This motivated research in Physics-Informed Neural Networks (PINNs) \cite{raissi2019physics}. The field has been growing at an exponential rate ever since its advent in 2019. PINNs can be used to solve a wide range of problems involving (partial) differential equations, namely forward, inverse, model discovery, surrogate modeling and uncertainty quantification. However, PINNs have an Achilles heel. Namely, how to balance the trade-off between fitting the data and respecting the laws of physics in the absence of validation data? We are therefore dealing with a multi-objective optimization problem. Here, ideas such as the ones proposed in \cite{kendall2018multi} for multi-task learning using uncertainty to weigh different objectives could be extended to solve our multi-objective optimization problem. Here, another challenge that we need to overcome is avoiding trivial solutions (e.g., any constant function is a solution to the Navier Stokes equations). A feasible strategy is to explicitly include a term in our training objective function that encourages non-trivial solutions.

\section{Concluding Remarks}
Artificial intelligence (AI) evangelizes the idea of automation. On the surface, AI algorithms take the data, develop their own understanding of it, and generate valuable insights and predictions -- all without human intervention. In truth, AI involves an enormous amount of repetitive manual operations, all hidden behind the scenes. This is what we call the ``iteration process''. Among many other degrees of freedom, this process entails model engineering (e.g., neural network architecture design) and management, experiment tracking, dataset versioning and augmentation. The iteration process is typically carried out by data engineers, data scientists, machine learning engineers, and other highly-trained (and highly-paid) specialists. However, at least part of their work can be streamlined by AutoML. In recent years, AutoML has demonstrated some promise in solving simple supervised learning problems, in particular (image) classification. However, this does not mean that AutoML will be successful in the face of more complex problems beyond (image) classification. It remains to be seen and tested in practice.

\newpage

% \section*{References}
\bibliographystyle{IEEEtran}
\bibliography{references.bib}

\end{document}